\documentclass[10pt,twocolumn,letterpaper]{article}

\usepackage{cvpr}              %

\usepackage[dvipsnames]{xcolor}

\definecolor{cvprblue}{rgb}{0.21,0.49,0.74}
\usepackage[pagebackref,breaklinks,colorlinks,citecolor=cvprblue]{hyperref}
\usepackage{multirow}

\usepackage{graphicx}
\usepackage{float}

\usepackage{lipsum}  
\usepackage{bbm}

\definecolor{mypurple}{RGB}{0, 128, 255}
\definecolor{dexcolor}{RGB}{0, 117, 4}
\definecolor{honotatecolor}{RGB}{1, 0, 253}

\newcommand\blfootnote[1]{%
  \begingroup
  \renewcommand\thefootnote{}\footnote{#1}%
  \addtocounter{footnote}{-1}%
  \endgroup
}

\newcommand{\colorRef}[1]{\textcolor{red}{#1}} %
\newcommand{\reffig}[1]{\colorRef{Fig.~\ref{#1}}}

\newcommand{\refFig}[1]{\mbox{\colorRef{Figure~\ref{#1}}}}
\newcommand{\reftab}[1]{\colorRef{Tab.~\ref{#1}}}
\newcommand{\refTab}[1]{\mbox{\colorRef{Table~\ref{#1}}}}
\newcommand{\refeq}[1]{\colorRef{Eq.~\ref{#1}}}

\newcommand{\refsec}[1]{\colorRef{Sec.~\ref{#1}}}

\newcommand{\subtitle}[1]{\textbf{#1}.}

\definecolor{GreenColor}{rgb}{0.137,0.573,0.565}
\definecolor{OrangeColor}{rgb}{0.914,0.541,0.0.141}
\definecolor{PurpleColor}{rgb}{0.5,0,0.7}
\definecolor{BlueColor}{rgb}{0,0.725,0.949}
\definecolor{PinkColor}{rgb}{0.9843,0.19215,0.6}

\newcommand{\M}[1]{\mathbf{#1}} %
\newcommand{\V}[1]{\mathbf{#1}} %
\newcommand{\R}{\rm I\!R}

\newcommand{\norm}[1]{\left\lVert#1\right\rVert}

\newcommand{\myparagraph}[1]{\noindent\textbf{#1:}}

\newcommand{\nameCOLOR}[1]{\textcolor{black}{#1}} %
\newcommand{\methodname}{\mbox{\nameCOLOR{HOLD}}\xspace}
\newcommand{\methodnet}{\mbox{\nameCOLOR{HOLD-Net}}\xspace}

\newcommand{\TITLE}{\methodname: Category-agnostic 3D Reconstruction of\\ Interacting Hands and Objects from Video}

\newcommand{\methodfullname}{Hand and Object reconstruction by Leveraging interaction constraints in three Dimensions}

\newcommand{\suppl}{\textcolor{black}{SupMat}\xspace}

\newcommand{\rgb}{RGB\xspace}

\newcommand{\sixD}{{6D}\xspace}

\newcommand{\threeD}{\xspace{3D}\xspace}

\newcommand{\sota}{{SOTA}\xspace}

\newcommand{\groundtruth}{{ground-truth}\xspace}

\newcommand{\point}{\mathbf{x}}
\newcommand{\texture}{\mathbf{c}}
\newcommand{\distance}{d}

\def\paperID{} %
\def\confName{CVPR}
\def\confYear{2024}

\title{\TITLE} %

\author{
Zicong Fan$^{1,2}$ 
\quad Maria Parelli$^{1}$ 
\quad Maria Eleni Kadoglou$^{1}$ 
\quad Muhammed Kocabas$^{1,2}$ \\
\quad Xu Chen$^{1,2,\dagger}$ 
\quad Michael J. Black$^{2}$ 
\quad Otmar Hilliges$^{1}$
 \\
 {
 $^1$ETH Z{\"u}rich, Switzerland \quad
 $^2$Max Planck Institute for Intelligent Systems, T{\"u}bingen, Germany
 }
}

\begin{document}

\twocolumn[{%
\renewcommand\twocolumn[1][]{#1}%
\maketitle
\begin{center}
    \centerline{\includegraphics[width=\linewidth]{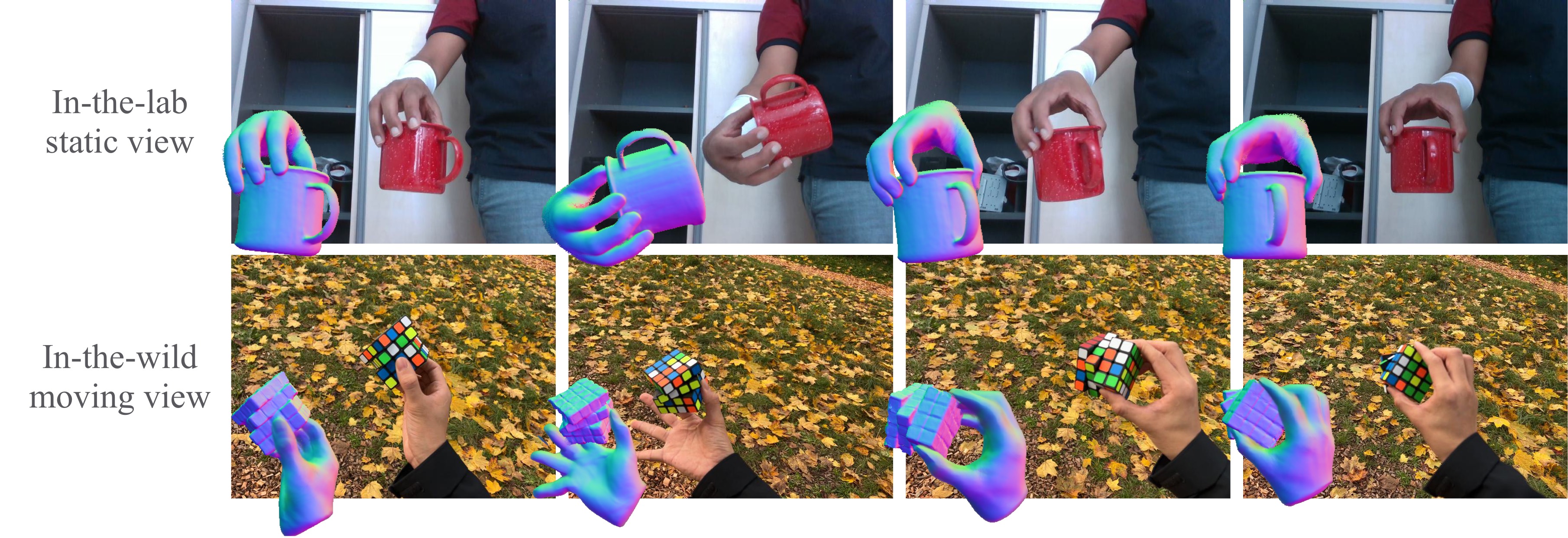}}
    \captionof{figure}{\textbf{HOLD:} Given a monocular video sequence of a hand interacting with an unknown object, our method, \methodname, reconstructs high-quality \threeD hand and object surfaces  in both in-the-lab videos from a static camera and in-the-wild egocentric-view videos.
    Here we show the input images and the reconstructed normals. Best viewed in color.
    }
    \label{fig:teaser}
\end{center}%

}
]
\blfootnote{$\dagger$ Work done prior joining Google}

\begin{abstract}

Since humans interact with diverse objects every day, the holistic \threeD capture of these interactions is important to understand and model human behaviour.
However, most existing methods for hand-object reconstruction from \rgb either assume pre-scanned object templates or heavily rely on limited \threeD hand-object data, restricting their ability to scale and generalize to more unconstrained interaction settings. 
To this end, we introduce \methodname\ -- the first category-agnostic method that reconstructs an articulated hand and object jointly from a monocular interaction video. We develop a compositional articulated implicit model that can reconstruct disentangled 3D hand and object from 2D images. We also further incorporate hand-object constraints to improve hand-object poses and consequently the reconstruction quality.
Our method does not rely on \threeD hand-object annotations while outperforming fully-supervised baselines in both in-the-lab and challenging in-the-wild settings.
Moreover, we qualitatively show its robustness in reconstructing from in-the-wild videos. 
Code: \url{https://github.com/zc-alexfan/hold}

\end{abstract}

\section{Introduction}
\label{sec:intro}

We interact with a diverse set of objects in our everyday lives: We hold our morning cup of coffee; we hold a drill when making home renovation; and we pour cereal from a box. 
Studies show that on average, we interact with 140 objects per day~\cite{perasso2015}.
To understand and model these interactions, it is critical to be able to reconstruct them in 3D.
Toward this goal, we tackle the challenging problem of reconstructing diverse 3D objects and the articulated hands holding them from only a monocular video of the hand-object interaction, as illustrated in \reffig{fig:teaser}.

Most hand-object reconstruction methods assume a pre-scanned object template~\cite{fan2023arctic,Hasson2020photometric,Yang_2021_CPF,hasson2021towards}, making it infeasible to scale to in-the-wild scenarios~\cite{cao2021handobject}. 
Other methods do not assume object templates~\cite{hasson2019obman,karunratanakul2020graspField,ye2022hand}, but are trained using datasets with a limited number of objects, leading to poor generalization. Very recently, Ye \etal~\cite{ye2023vhoi} introduced a data-driven prior that is trained on six object categories and they leverage this prior to reconstruct hand and object surfaces from segmentation mask observations. 
Although they can reconstruct novel objects and articulated hands, their method is limited to these training categories. 
Another emerging line of work focuses on in-hand object scanning~\cite{hampali2023inhand,zhong2024colorneus,huang2022reconstructing} from monocular videos. They adapt multi-view reconstruction techniques to aggregate observations of hand-held objects in multiple rigid poses. While achieving promising reconstruction quality on novel objects, these methods do not consider hand articulation and hence cannot handle more dexterous hand-object interaction.

In this paper, we go beyond prior works to tackle the problem of \textit{category-agnostic reconstruction} of hands and objects.
Given a monocular video as input, our method \methodname (\methodfullname) reconstructs   hand and object \threeD surfaces for every frame without assuming an object template.
Our key insight is that hands and objects in interaction provide complementary cues to each other's shapes and poses. 
For example, when one holds a coffee mug, the hand geometry infers the shape of the mug via contact. Thus, we propose to jointly model the object and articulated hand with a compositional neural implicit model.

To jointly reconstruct the hand and object surfaces from a video,
\methodname performs initial hand pose estimation via an off-the-shelf hand regressor and object pose estimation with structure-from-motion (SfM). 
With the initial noisy hand and object poses, we train \methodnet, our compositional neural implicit model of an articulated hand, and an object.
The model is volumetrically rendered and supervised with auxiliary losses to obtain the \threeD hand and object surfaces. 
After initializing the hand and object shapes by training \methodnet, we optimize hand and object poses via interaction constraints. 
Finally, we use the refined poses to train \methodnet for better shape reconstruction.

We empirically show that by jointly modelling the hand and object in this category-agnostic reconstruction setting through interaction constraints, we achieve better reconstruction quality than methods that only consider objects. We quantitatively evaluate our method with an existing hand-object dataset and further show that our method can generalize to  both in-the-lab and in-the-wild videos.
We also demonstrate generalization to videos captured by a moving camera from both 3rd person and 1st person views with diverse lighting and background conditions.

To summarize our contributions: 
1) We present a novel method that accurately reconstructs \threeD hand and object surfaces from monocular 2D interaction videos without requiring a pre-scanned object template or pre-trained object categories. 
2) We formulate a compositional implicit model that facilitates the disentanglement and the reconstruction of 3D hand and object.
3) We show that by jointly optimizing hand-object constraints, we can obtain better reconstruction quality than treating the hand and object separately. 
4) We evaluate our method both qualitatively and quantitatively for \threeD reconstruction, and we  demonstrate realistic reconstruction on challenging in-the-wild videos.
Our model and code will be available for research.

\begin{figure*}[t]
    \centerline{\includegraphics[width=1\linewidth, trim=400 700 400 0, clip]{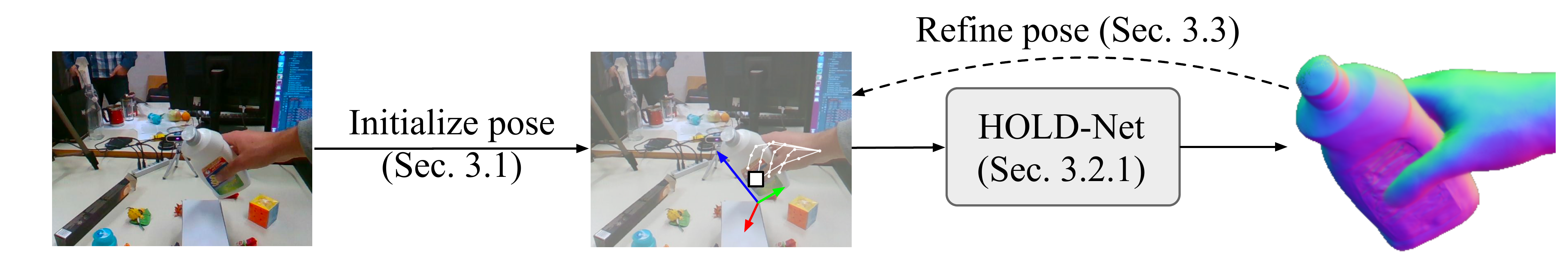}}
    \caption{
        \subtitle{Method overview}
        For each image in a video,
        our method, \methodname, first initializes the hand and object poses using off-the-shelf estimators. Then we briefly pre-train \methodnet, a compositional implicit signed distance field to learn hand and object shapes. The learned shapes of \methodnet are then used to refine poses with hand-object interaction constraints. Finally, we use the refined poses to fully train \methodnet to learn accurate \threeD geometries of hand and object.
    }
    \label{fig:method}
\end{figure*}
\section{Related Work}
\label{sec:related_work}
\myparagraph{\threeD hand pose and shape recovery}
The field of monocular RGB 3D hand reconstruction has been evolving since the foundational work of Rehg and Kanade~\cite{Rehg:1994}. A significant portion of the existing literature is focused exclusively  on reconstructing the hand~\cite{iqbal2018hand,Mueller2018ganerated,Spurr2018crossmodal,spurr2021peclr,spurr2020eccv,zimmermann2017iccv,Boukhayma2019,hasson2019obman,ziani2022tempclr,fan2021digit,simon2017hand,Zhang2019endtoend,interhand,li2022interacting,zhang2021interacting,tzionas2013directional,chen2023handavatar,fu2023deformer}. For instance, Zimmermann \etal~\cite{zimmermann2017iccv} employ a deep convolutional network, implementing a multi-stage approach to achieve \threeD hand pose estimation. 
Spurr \etal~\cite{spurr2020eccv} incorporate biomechanical constraints to refine and stabilize the predictions of hand poses.
Recently, there are also methods that reconstruct \threeD hand poses of strongly interacting hands~\cite{moon2023bringing,lee2023im2hands,li2022interacting,meng20223d,interhand,moon2023dataset,li2023renderih,ohkawa2023assemblyhands,guo2023handnerf}. 
Tse \etal~\cite{tse2023spectral} introduce a spectral graph-based transformer for two hand reconstruction.
Compare to these methods, we focus on hand-object reconstruction. 

\myparagraph{Hand-object reconstruction} 
Reconstructing the hand and object in \threeD from images and videos is also a well-established research area~\cite{hasson2019obman,liu2021semi,Yang_2021_CPF,grady2021contactopt,Hasson2020photometric,tekin2019ho,corona2020ganhand,zhou2020monocular,hasson2021towards,tse2022collaborative,yang2021cpf}.
Most methods in the literature assume an object template and only estimate the object and hand pose  \cite{tekin2019ho,corona2020ganhand,liu2021semi,cao2021handobject,yang2021cpf}.
For example, 
Tekin \etal~\cite{yang2021cpf} infer \threeD control points for both the hand and the object in videos, using a temporal model to propagate information across time. 
Liu \etal~\cite{liu2021semi} devise a semi-supervised learning approach by first constructing pseudo-groundtruth on hand-object interaction videos based on temporal heuristics and train the model with the new annotation.
Yang \etal~\cite{yang2021cpf} introduce a contact potential field for better hand-object contact for a given object.
Despite accurate object pose estimation quality, it is  hard to generalize such work to novel objects and in-the-wild videos because it requires known object templates.
There are methods that do not assume an object template by training on \threeD hand-object  data~\cite{karunratanakul2020graspField,hasson2019obman,ye2022hand,chen2023gsdf}.
Unfortunately, these methods have poor generalization ability due to limited \threeD hand-object data.
Recently, there are more generalizable approaches~\cite{swamy2023showme,Prakash2023ARXIV,qu2023novel,ye2023vhoi,huang2022reconstructing} with differentiable rendering and data-driven priors.
However, they require either the hand to be rigid when interacting with objects~\cite{huang2022reconstructing,Prakash2023ARXIV}, multi-view observation~\cite{qu2023novel}, or category-level hand-object supervision~\cite{ye2023vhoi}.
In contrast to them, our method allows articulated hands, only requires monocular view, and is category-agnostic.

\myparagraph{In-hand object scanning} 
There have been increasing interests in in-hand object scanning.
The goal of this task is to reconstruct canonical \threeD object shape from a video of a human interacting with an object; the hand is often not reconstructed.
For example, early works such as Tzionas \etal~\cite{tzionas20153d} leverage hand motion as a prior for object scanning. 
Recently, BundleSDF~\cite{wen2023bundlesdf}  estimates the object pose with the help of sequential RGBD images and simultaneously reconstructs the implicit surface defined by the Signed Distance Function (SDF). 
HHOR~\cite{huang2022reconstructing} also employs SDFs for object surface representation but distinguishes itself by concurrently reconstructing both the object and the hand, assuming the object is securely gripped.
Hampali \etal~\cite{hampali2023inhand} propose a novel approach, incorporating a camera trajectory alignment technique and utilizing volumetric rendering for enhanced object surface reconstruction.
Very recently, Zhong \etal~\cite{zhong2024colorneus} introduce a global coloring and relighting network that significantly improves texture extraction during the object scanning process.
In contrast to our work, the methods above do not reconstruct hands with articulation and mainly focus on capturing the object canonical shape.

\section{Method: \methodname}
\label{sec:method}
\refFig{fig:method} summarizes our method, \methodname, for reconstructing hand-object surfaces from a monocular \rgb video. 
To achieve this, \methodname first initializes hand and object poses (\refsec{sec:pose_init}) for each frame in a video. 
Then we use the poses to train \methodnet (\refsec{sec:pretraining}), a compositional implicit signed distance field for hand and object shapes with a small amount of epochs. 
Using the learned shapes, we refine hand-object poses via interaction constraints (\refsec{sec:pose_refine}). 
Finally, with the refined poses we fully train \methodnet (\refsec{sec:final_training}), resulting in accurate \threeD hand-object geometry.

\subsection{Pose initialization}
\label{sec:pose_init}
For each frame, to obtain hand poses $\theta$, shape $\beta$, global rotation $\V{R}_h\in SO(3)$ and translation $\V{t}_h\in \R^3$, we use an off-the-shelf hand pose estimator~\cite{lin2021end-to-end}. 
Estimating object pose is more challenging because our approach is category agnostic and 
existing category-level object pose estimators are unsuitable for out-of-category objects~\cite{chen2020category,wang2019normalized}. 
Consequently, following~\cite{hampali2023inhand}, we first create object-only images for each video by masking out the object pixels using an off-the-shelf segmentation network~\cite{kirillov2023segment}.
We then use HLoc~\cite{sarlin2019coarse,sarlin2020superglue} to perform structure-from-motion (SfM) to obtain a point cloud defining the object and its rotation $\V{R}_o\in SO(3)$ and translation $\V{t}_o\in \R^3$ for each frame.
Since SfM only reconstructs point clouds up to a scale, to align the hand and object in the same space and to estimate the object scale $s\in \R$, we perform a simple optimization procedure that encourages hand-object contact while enforcing the 2D reprojection of hand joints and the object point cloud to match with the original 2D projection. 
This optimization updates hand and object translation $\{\V{t}_h, \V{t}_o\}$ for each frame, the hand shape $\beta$, and an object scale $s$.
Details in \suppl.

\subsection{\methodnet training}
\label{sec:pretraining}
\subsubsection{\methodnet}
\begin{figure*}[t]
    \centerline{\includegraphics[width=1\linewidth, trim=0 0 0 340, clip]{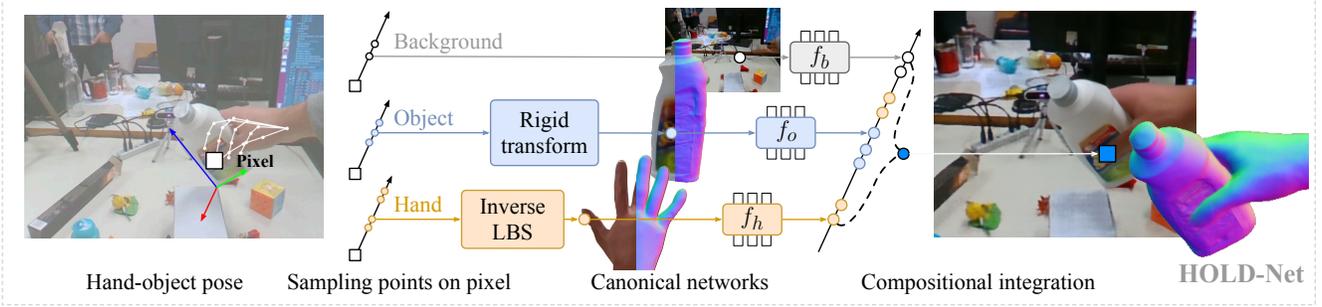}}
    \caption{
        \subtitle{\methodnet} Given as input hand and object poses and a query pixel, \methodnet determines the pixel color in the following steps. 1) \methodnet first samples points along the ray independently for object, hand, and background using error-bounded sampling. 2) These sampled points in the observation space are then mapped to the canonical space. Object points are rigidly transformed based on the object pose and hand points undergo articulated deformation with inverse linear blend skinning. 3) The SDF and color values for the sampled points are queried from the canonical hand, object, and background networks. 4) All object,  hand, and background points are aggregated and their color and density values are integrated to determine the pixel color.     
    }
    \label{fig:holdnet}
\end{figure*}

Inspired by~\cite{guo2023vid2avatar,yariv2021volume}, we represent the hand and object surfaces as two neural representations that can be volumetrically rendered into an \rgb image. We use a time-dependent NeRF++~\cite{zhang2020nerf++} to model the dynamic background. Our \methodnet model is illustrated in \reffig{fig:holdnet}

\myparagraph{Hand model}
We model the hand as an implicit network, driven by MANO pose $\theta$, global rotation $\V{R}_h$, and translation $\V{t}_h$.
To model the hand shape and appearance in canonical space, we use a signed distance and texture field  parameterized by a multi-layer perception (MLP):
\begin{align}
    f_h: \mathbb{R}^{3}  &\rightarrow \mathbb{R} \times \mathbb{R}^{3} \\
         \point &\mapsto \distance, \texture,
\end{align}
where the MLP $f_h$, with learnable parameters $\psi_h$, takes in a canonical point $\point$, and predicts its signed distance values to the hand surface $\distance$ and color $\texture$.

To determine the signed distance and color in the deformed observation space, we map points in the observation space $\mathbf{x}'$ back to the canonical space using inverse LBS:
\begin{align}
    \point = (\textstyle{\sum_{i=1}^{n_b}} w_i(\point') \cdot \mathbf{B}_i)^{-1} \point',
    \label{eq:inverse}
\end{align}
where $\{\mathbf{B}_i\}_{i=1,\ldots,n_b}$ are the bone transformations derived from  $\theta$ with forward kinematics, and $\{w_i(\point')\}_{i=1,\ldots,n_b}$ are the skinning weights of each deformed point determined by averaging the skinning weights of the K-nearest vertices of the MANO model~\cite{mano} weighted by the distance.

\myparagraph{Object model}
Similar to the hand model, our object model is driven by
the relative object scale $s$, rotation $\V{R}_o$ and translation $\V{t}_o$ between the canonical and deformed space respectively.
The object canonical shape and texture are modelled via a neural signed distance and texture field $f_o$, with learnable parameters $\psi_o$:
\begin{align}
    f_o: \mathbb{R}^{3} \times \mathbb{R}^{n_{z,o}} &\rightarrow \mathbb{R} \times \mathbb{R}^{3} \\
         \point, \mathbf{z_o} &\mapsto \distance, \texture,
\end{align}
where $\mathbf{z_o}\in\mathbb{R}^{n_{z,o}}$ of dimension $n_{z,o}=32$ is an optimizable dependent latent code to model the changing object appearance due to varying pose, occlusion and shadows.

To determine the signed distance and color of the object in the deformed observation space, we map points in the observation space $\mathbf{x}'$ back to the canonical space using a simple rigid transformation:
\begin{align}
    \point = (s\V{R}_o)^{-1} \cdot (\point' - \V{t}_o).
    \label{eq:inverse}
\end{align}

\myparagraph{Background}
Following \cite{yariv2021volume,guo2023vid2avatar}, we define a bounding sphere of the foreground scene, in our case the hand and the object. For a given sample $\point'$ outside the bounding sphere, the signed distance and color are predicted by a background network with learnable parameters $\psi_b$:
\begin{align}
    f_b: \mathbb{R}^{3} \times \mathbb{R}^{3} \times \mathbb{R}^{n_{z,b}}  &\rightarrow \mathbb{R} \times \mathbb{R}^{3} \\
         \point, \mathbf{v}, \mathbf{z_b} &\mapsto \distance, \texture,
\end{align}
where $\mathbf{v} \in \mathbb{R}^{3}$ is the viewing direction and $\mathbf{z}\in\mathbb{R}^{n_{z,b}}$ of dimension $n_{z,b}=32$ is an optimizable latent code with distinct value for each frame to model dynamic backgrounds. 
Since we are only interested in modelling hands and objects, and images of human interaction often obtain parts of the body, this model is also used to explain partial observation of the human body as part of the changing background.
Following NeRF++~\cite{zhang2020nerf++}, we use their inverted sphere parametrization in our background model. 
More details can be found in \suppl.

\myparagraph{Compositional volumetric rendering}
Following~\cite{yariv2021volume}, to convert hand and object SDFs to density $\sigma$ for volume rendering, we use the cumulative distribution function of the scaled Laplace distribution, denoted as $\Gamma_{\alpha_1, \alpha_2}(s)$, where $\alpha_1, \alpha_2 >0$ are optimizable  (see~\cite{yariv2021volume} for details).

To render the foreground, \ie, the hand and object, we first sample points along the corresponding ray $\V{r}$ parameterized by a camera center $\V{o}$ and a viewing direction $\V{v}$ using error-bounded sampling~\cite{yariv2021volume}. We sample $n$ points for the hand $\{\V{x}'\}_{i=1, \ldots,n}^h$, transform them to canonical space using inverse LBS, and query their opacity and color values $\{(\sigma_{i}, \V{c}_{i})\}^h_{i=1, \ldots,n}$ from the canonical hand model $f_h$.
Similarly for the object, we sample $n$ points $\{\V{x}_{i}\}_{i=1, \cdots,n}$ along the same ray, and obtain their density and color $\{(\sigma_{i}, \V{c}_{i})\}^o_{i=1, \ldots,n}$ by transforming them rigidly back to the canonical object model.
We then sort and merge the two sets of samples via their depth value to obtain $\{(\sigma_{i}, \V{c}_{i})\}_{i=1, \ldots,2n}$  and perform volumetric rendering:
\begin{align}
    &C_F(\mathbf{r})=\textstyle{\sum_{i=1}^{2n}}\tau_i \V{c}_{i} \label{eq:fg_render}\\
&\text{where }
    \tau_i=\exp \left(-\textstyle{\sum_{j<i}} \sigma_{j} \delta_j\right)\left(1-\exp \left(-\sigma_{i} \delta_i\right)\right) \notag 
\end{align}
\noindent and $\delta_i$ is the distance between two consecutive samples. Similarly, we  determine the background color  $C_B(\V{r})$ by querying the density and color of sampled points from the background network. To composite the background and foreground, we render the foreground mask probability of a ray $\V{r}$, which can be derived as $M_F(\V{r}) = \sum_{i=1}^{2n} \tau_i \in\R$. To render with the dynamic background, the final color value of the ray is defined as
 \begin{align}
C(\V{r}) = C_F(\V{r}) + (1-M_F(\V{r}))C_B{(\V{r})}
\label{eq:composition}
\end{align}
where $C_B(\V{r})$ is the background color value. Similar to determining the foreground probability, our model also determines the amodal mask~\cite{li2016amodal} probability of a pixel belonging to hand $M_h(\V{r})\in\R$ or object $M_o(\V{r})\in\R$ by accumulating the transmittance of  hand or object samples independently. In addition, our model renders the class probability $S(\V{r})\in\R^3$ between hand, object, and background of each pixel by following the rendering procedure in \refeq{eq:fg_render} and \refeq{eq:composition}, while replacing the color $\V{c}$ of each sample point with a one-hot three-vector for each class.

\subsubsection{Training losses} 
Since reconstructing the \threeD hand and object shapes from a monocular video is highly under-constrained, we devise a loss $\mathcal{L}$ consisting of several terms to optimize for the texture and shape network parameters $\{\psi_h, \psi_o, \psi_b\}$, the per-frame parameters $\{\theta, \V{R}_h, \V{t}_h, \V{R}_o, \V{t}_o, \mathbf{z_o}, \mathbf{z_b}\}$, and global parameters $\{\beta, s\}$. 

In particular, we first encourage \rgb values to be consistent with the input image via 
\begin{align}
\mathcal{L}_\text{rgb} =  \sum_{\textbf{r}} \norm{C(\V{r}) - \hat{C}(\V{r})}
\end{align}
where $\V{r}$ is a ray casted from a sampled pixel on an image, and $C(\V{r})$ and $\hat{C}(\V{r})$ are the rendered and \groundtruth color.

To encourage disentanglement between the hand, object, and background, we enforce a multi-class segmentation loss 
\begin{align}
    \mathcal{L}_\text{segm} = \sum_{\textbf{r}} \norm{S(\V{r}) - \hat{S}(\V{r})},
\end{align}
where $\hat{S}(\V{r}) \in \R^3$ is a one-hot vector representing the predicted class of a pixel, obtained with an off-the-shelf segmentation network~\cite{kirillov2023segment}.
To regularize the hand and object shapes, we sample points uniformly at random as well as around the surface of the hand and object in their canonical space.
We then enforce the eikonal loss $\mathcal{L}_\text{eikonal}$~\cite{gropp2020igr}  to regularize the canonical hand and object shapes. 
To provide a shape prior for the hand, using the same set of samples, we enforce the SDF predicted by our canonical hand model to be similar to the one from a MANO model using the following loss:
\begin{align}
    \mathcal{L}_\text{sdf} = \sum_{\point\in 
\mathcal{X}} \norm{f_h(\point) - SDF(\point)}
\end{align}
where $\mathcal{X}$ is a set of randomly sampled points in canonical space and $SDF(\point)$ is the signed distance from the MANO mesh. To obtain a smooth SDF from the MANO mesh, we sub-divide MANO using Loop subdivision~\cite{loop1987smooth}.

Finally, to enforce sparsity of the hand density outside of its surface, for a ray $\V{r}$ that is far from the MANO hand mesh, we enforce its amodal mask probability $M_h\V{(r)}$ to be zero. 
A ray $\V{r}$ is far away from a mesh if its closest distance to the mesh exceeds a threshold.
Similarly, we periodically construct an object mesh via marching cubes and use it to enforce the object sparsity loss when the ray of a pixel is far away from the object.
Formally,
\begin{align}
    \mathcal{L}_\text{sparse} = \sum_{\textbf{r}\in \mathcal{F}_h} \norm{M_h(\V{r})} + \sum_{\textbf{r}\in \mathcal{F}_o} \norm{M_o(\V{r})}
\end{align}
where $\mathcal{F}_h$ and $\mathcal{F}_o$ are the set of rays that are far from the hand and object meshes respectively. 
The total loss $\mathcal{L}$ is defined as  
\begin{align}
\mathcal{L} =
     \mathcal{L}_\text{rgb} &+ \lambda_\text{segm}\mathcal{L}_\text{segm} + \lambda_\text{sdf}\mathcal{L}_\text{sdf} \notag \\ 
     &+ \lambda_\text{sparse}\mathcal{L}_\text{sparse} +
    \lambda_\text{eikonal}\mathcal{L}_\text{eikonal} 
\end{align}
where $\lambda_{*}$ are the weights for the losses. 
Note that since predicted segmentation masks are often noisy we gradually decrease $\lambda_\text{segm}$ over time and gradually increase the prior weights $\lambda_\text{sdf}$ and $\lambda_\text{sparse}$ over time.

\subsection{Pose refinement}
\label{sec:pose_refine}
The poses from \refsec{sec:pose_init} are imperfect because object point clouds from SfM are noisy, and the hand shape parameters are not optimized. 
While jointly training \methodnet and optimizing the poses could theoretically resolve noisy poses, 
we empirically find that this strategy is inefficient as the pose of each training frame gets only sparse training signals, i.e.~only when the the corresponding frame is sampled.
To obtain accurate poses efficiently, we first train \methodnet for a small number of epochs to obtain a coarse estimate of the object shape. 
Then we follow ~\cite{yin2023hi4d} and refine the hand and object pose parameters $\{\V{R}_h, \V{t}_h, \V{R}_o, \V{t}_o, \beta, s\}$ with mesh-based constraints, using the object mesh extracted from \methodname and MANO.

In particular, we encourage contact between frequently contacted hand vertices $\M{V}_{tips}$ (vertex ids from~\cite{hasson2019obman}) and the object vertices by encouraging each such hand vertex to be close to an object vertex. 
Formally, the loss is defined as:
\begin{align}
    \mathcal{L}_\text{contact} = \sum_i \min_j \norm{\M{V}^i_\text{tips} - \M{V}^j_o}.
\end{align}
To provide better pixel-alignment for the hand and the object, we use Soft Rasterizer~\cite{liu2019soft} to render the hand amodal masks $\mathcal{M}_h$ and object amodal masks $\mathcal{M}_o$ and encourage it to match the masks from off-the-shelf semantic segmentation using an occlusion-aware term $\mathcal{L}_\text{mask}$ similar to~\cite{zhang2020perceiving} (see \suppl). 
These simple terms work well in practice; see \suppl for more discussion.

\subsection{Final training} 
\label{sec:final_training}
Given the refined hand pose parameters $\{\theta\}$ from \refsec{sec:pretraining} and $\{\beta, \V{R}_h, \V{t}_h, \V{R}_o, \V{t}_o, s\}$ from \refsec{sec:pose_refine},
we fully train \methodnet with the loss $\mathcal{L}$  following the formulation in \refsec{sec:pretraining} to reconstruct the \threeD hand and object geometries for every frame of an input video.
To avoid artifacts that $f_h$, $f_o$, $f_b$ could have learnt during pre-training due to inaccurate poses, we train $\{\psi_h, \psi_o, \psi_b, \V{z}_o, \V{z}_b\}$ from scratch.
For brevity, we ignore the timestamp for frame-specific parameters.
Note that \methodnet is pre-trained with half the number of epochs compared to this full-training stage for computational efficiency as we observe that the hand and object shape stabilizes in the early stages of training.

\section{Experiments}
\label{sec:exp}
\begin{figure*}[t]
    \centerline{\includegraphics[width=1.0\linewidth]{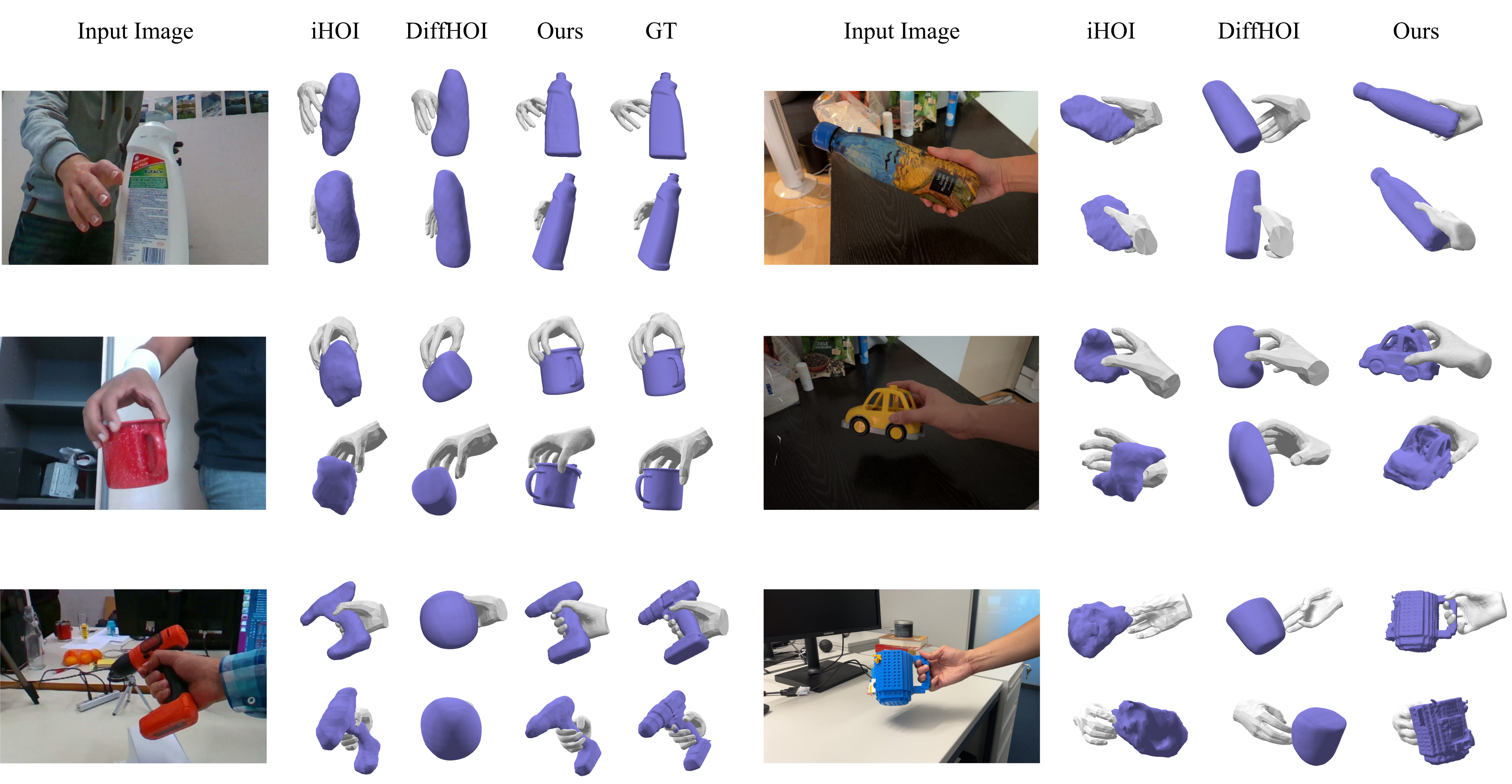}}
    \caption{
        \subtitle{Qualitative comparison with \sota} We show hands and objects reconstructed by our method and \sota baselines from in-the-lab (\emph{left}) and in-the-wild (\emph{right}) videos. Our reconstruction demonstrates more accurate shapes, richer details, and more accurate poses. In addition, our method works consistently well on various objects, even those with unique shapes (e.g.~the Lego mug on the bottom right).
    }
    \label{fig:quali_sota}
\end{figure*}%

In this section, we compare our method with existing baselines for the \textit{category-agnostic hand-object reconstruction} task.
The goal of the task is to reconstruct accurate \threeD surfaces for the hand and object from monocular video observations, where we do not assume an object template.

\myparagraph{In-the-lab dataset}
We use the widely-used HO3D-v3 dataset~\cite{hampali2020honnotate} for quantitative and qualitative evaluation. The dataset consists of \rgb videos of a human hand manipulating a rigid object. The hand is articulated in this dataset and it provides accurate \threeD annotations for MANO hand parameters and  \sixD object poses. Since the dataset does not release \groundtruth for its evaluation set, we use sequences in the training set for evaluation.
Since our task focuses on generalization, none of the methods here are trained using HO3D.
We downsample all sequences every 5 frames.

\myparagraph{In-the-wild sequences}
To evaluate whether our method can generalize to diverse in-the-wild settings, we capture sequences of household items in both in-door and out-door scenes. We capture in both 1st-person moving view and 3rd-person static view via an iPhone 14 main camera under different lighting conditions. %
For each video, we downsample it every 10 frames for our experiments.

\myparagraph{Metrics}
We use root-relative mean-per-joint error (MPJPE) in millimeters to measure hand pose error, and Chamfer distance in squared centimeters to evaluate object reconstruction quality~\cite{chen2023gsdf}. 
Since Chamfer distance is sensitive to outliers, we also use F-score in percentage to measure local shape details~\cite{ye2023vhoi,tatarchenko2019single}. 
In particular, to evaluate object template quality independent from object pose, 
following~\cite{ye2023vhoi}, we perform ICP alignment to the \groundtruth mesh of the HO3D meshes allowing scale, rotation and translation and compute the Chamfer distance (CD)  and F-score at 5mm (F5) and 10mm (F10).
To measure object pose and shape relative to the hand in \threeD, we subtract each object mesh by the predicted hand root and compute the hand-relative Chamfer distance for the object (CD$_h$).

\myparagraph{Implementation details}
We train each sequence using Adam~\cite{kingma2014adam}. In each iteration we 
 optimize 10 randomly sampled images from the sequence. 
For each image, we sample 256 pixels and for each pixel, we sample 64 points along the ray.
For stability, we perform gradient clipping.
We perform the initial training for 100 epochs, which requires around 10 hours using an A100 GPU.
The final training takes 200 epochs.
We use SAM-track~\cite{cheng2023segment} to derive the hand and object segmentation masks by using point-prompting for the first frame of each video. 
More details can be found in \suppl.

\begin{table}[t]
\resizebox{1.00\linewidth}{!}{

\begin{tabular}{cccccc}
\toprule
       & MPJPE [mm] $\downarrow$ & CD [cm$^2$] $\downarrow$ & F10 [$\%$]  $\uparrow$  & CD$_h$ [cm$^2$] $\downarrow$
\\
\midrule
iHOI$^\dagger$~\cite{ye2022hand}          & 38.4       & 3.8     & 75.8 & 41.7          \\
DiffHOI~\cite{ye2023vhoi}       & 32.3       & 4.3     & 68.8 & 43.8          \\
Ours          & \textbf{24.2}       & \textbf{0.4}     & \textbf{96.5} & \textbf{11.3}        
\\ \bottomrule
\end{tabular}
}
\caption{
\subtitle{Comparison with \sota hand-object reconstruction methods} We evaluate our method and two baselines on the HO3D dataset. 
\emph{$\dagger$ During training, iHOI uses 3D annotation of the test objects, while DiffHOI and ours do not use such information.}
}
\label{tab:sota_ho}
\end{table}

\subsection{State-of-the-art comparison}
\myparagraph{Hand-object reconstruction}
\refTab{tab:sota_ho} compares our method with existing hand-object reconstruction methods that do not assume an object template. 
We observe that \methodname  
significantly outperforms existing methods in terms of hand pose (MPJPE), object pose and shape (CD) accuracy. 
Our method also infers the relative spatial arrangement of the hand and object more accurately as shown by the superior hand-relative Chamfer distance (CD$_h$).

\begin{figure}[t]
    \centerline{\includegraphics[width=0.78\linewidth]{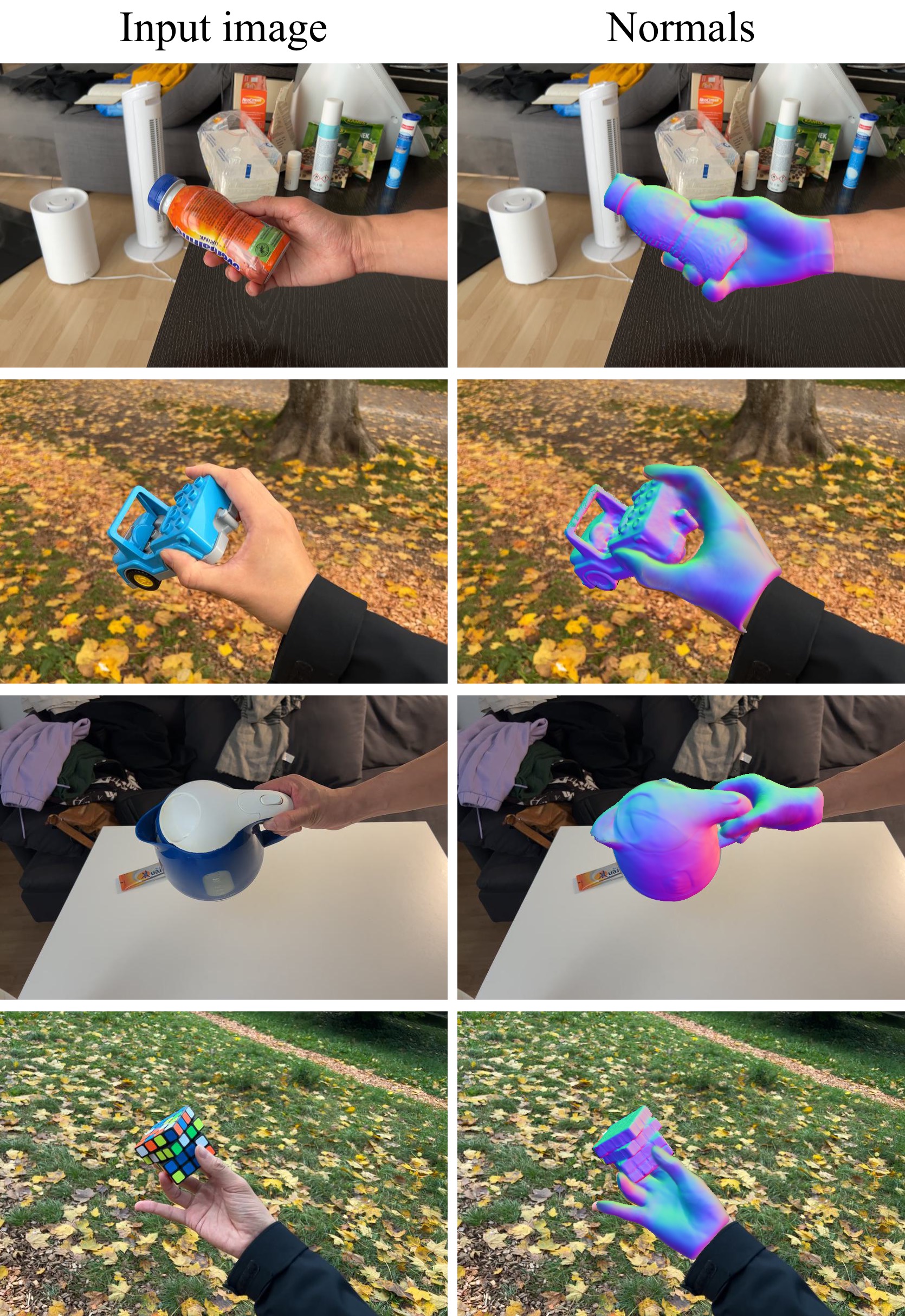}}
    \caption{
        \subtitle{More qualitative results} We render the normals of hands and objects reconstructed by our method. Our method can reliably reconstruct the objects in both static views and moving egocentric views.
    }
    \label{fig:more_quali}
    \vspace{-4mm}
\end{figure}

This improvement is also reflected in the qualitative comparison in \reffig{fig:quali_sota}. Our method consistently produces reconstructions that are closer to the ground-truth than those of iHOI and DiffHOI, with notable improvements in capturing the fine structures, such as the mug handle and the car frame, as well as the dynamic postures of the hand and object. In contrast, the reconstructions from the two baseline methods lack details and suffer from erroneous hand and object poses, even on the easier in-the-lab dataset. Notably, both baseline methods use 3D supervision - iHOI is trained on the HO3D dataset sequences with ground-truth \threeD shape and DiffHOI uses 3D shapes of diverse bottles and mugs as training supervision. In contrast, our method only uses the input 2D monocular video without requiring any 3D annotation, while still achieving superior  quality.
Our method can also reconstruct hands and objects reliably under different backgrounds, and lighting conditions in both  3rd-person view and  moving egocentric views (see~\reffig{fig:more_quali}). 

\myparagraph{Generalization} To quantify our method's ability to generalize compared to DiffHOI, in \refTab{tab:generalize} we split the HO3D sequences according to whether they belong to the training categories of DiffHOI. 
We see that while DiffHOI's performance significantly drops  across all metrics for unseen categories, our method has consistent performance on all categories. This is also reflected in \reffig{fig:quali_sota} by the samples on the right: our method can accurately reconstruct objects such as the drill, while the baseline methods do not generalize to  instances that are outside their training distributions.

\begin{table}[]
\resizebox{1.00\linewidth}{!}{
\begin{tabular}{ccccccc}
\toprule
                     & Object categories & MPJPE [mm] $\downarrow$ & CD [cm$^2$] $\downarrow$  & F10 [$\%$] $\uparrow$        & CD$_h$ [cm$^2$]  $\downarrow$  \\
                     \midrule
DiffHOI &\multirow{2}{*}{DiffHOI training}  & 34.2&1.3&83.5&42.5\\
                     Ours&     &\textbf{22.5}&\textbf{0.4}&\textbf{95.9}&\textbf{10.4}\\
                     \midrule
DiffHOI &\multirow{2}{*}{DiffHOI unseen} & 30.9&6.5&57.8&44.8\\
                     Ours&     &\textbf{25.5}&\textbf{0.3}&\textbf{96.9}&\textbf{12.0} \\
\bottomrule
\end{tabular}
}
\caption{
\subtitle{Generalization comparison} We compare the generalization ability of our method and the \sota method DiffHOI. We report results on objects within and beyond DiffHOI's training categories. DiffHOI's performance degrades significantly on unseen object categories while our method produces more accurate reconstruction consistently. 
}
\label{tab:generalize}
\end{table}

\begin{table}[t]
\centering
\resizebox{0.8\linewidth}{!}{
\begin{tabular}{cccc}
\toprule
        & CD [cm$^2$] $\downarrow$  & F5 [$\%$] $\uparrow$        & F10 [$\%$] $\uparrow$  \\
\midrule
Hampali~\cite{hampali2023inhand} & 1.4 & 57.4 & 79.9 \\
Ours    & \textbf{0.5} & \textbf{84.3} & \textbf{94.4} \\
\bottomrule
\end{tabular}
}
\caption{
\subtitle{Comparison with a \sota in-hand scanning method}
We compare our method with Hampali \etal~\cite{hampali2023inhand} following their protocol on HO3D.
}
\label{tab:inhand}
\end{table}

\begin{table}
\resizebox{1.00\linewidth}{!}{
\begin{tabular}{cccccccc}
\toprule
          & MPJPE [mm] $\downarrow$ & CD [cm$^2$] $\downarrow$ & F10 [$\%$] $\uparrow$ & CD$_h$ [cm$^2$] $\downarrow$
\\
\midrule
w/o hand        & -       & 0.41        & 95.9           & -         \\
w/o pose ref. & 24.6       & 0.55        & 94.2          & 122.1           \\
Ours       & \textbf{24.2}       & \textbf{0.38}        & \textbf{96.5}            & \textbf{11.3}          \\

\bottomrule
\end{tabular}
}

\caption{\subtitle{Ablation study} Modelling the hand and object jointly improves object reconstruction accuracy. Pose refinement improves object and hand poses and consequently object reconstruction accuracy. }
\label{tab:ablation}
\end{table}

Interestingly, our method significantly outperforms DiffHOI even for its training categories.
We can gain insight into this from the water bottle example on the top-right in \reffig{fig:quali_sota}: DiffHOI tries to reconstruct bottles seen in their data-driven prior training set, which leads to a generic bottle. In comparison, our reconstructed bottle realistically captures the shape details of  the one in the image.

\myparagraph{In-hand object scanning}
We also compare with the \sota method for in-hand object scanning from Hampali \etal~\cite{hampali2023inhand}; see \refTab{tab:inhand}.
Since there is no code released, we train our model following their selection of sequences and compare our canonical shape with the \threeD object results downloaded from their official webpage. We observe that our method recovers significantly better object canonical shapes (see CD) and local details (F5 and F10).
We refer to \suppl for qualitative comparison.

\subsection{Ablation}

\begin{figure}[t]
    \centerline{\includegraphics[width=0.8\linewidth]{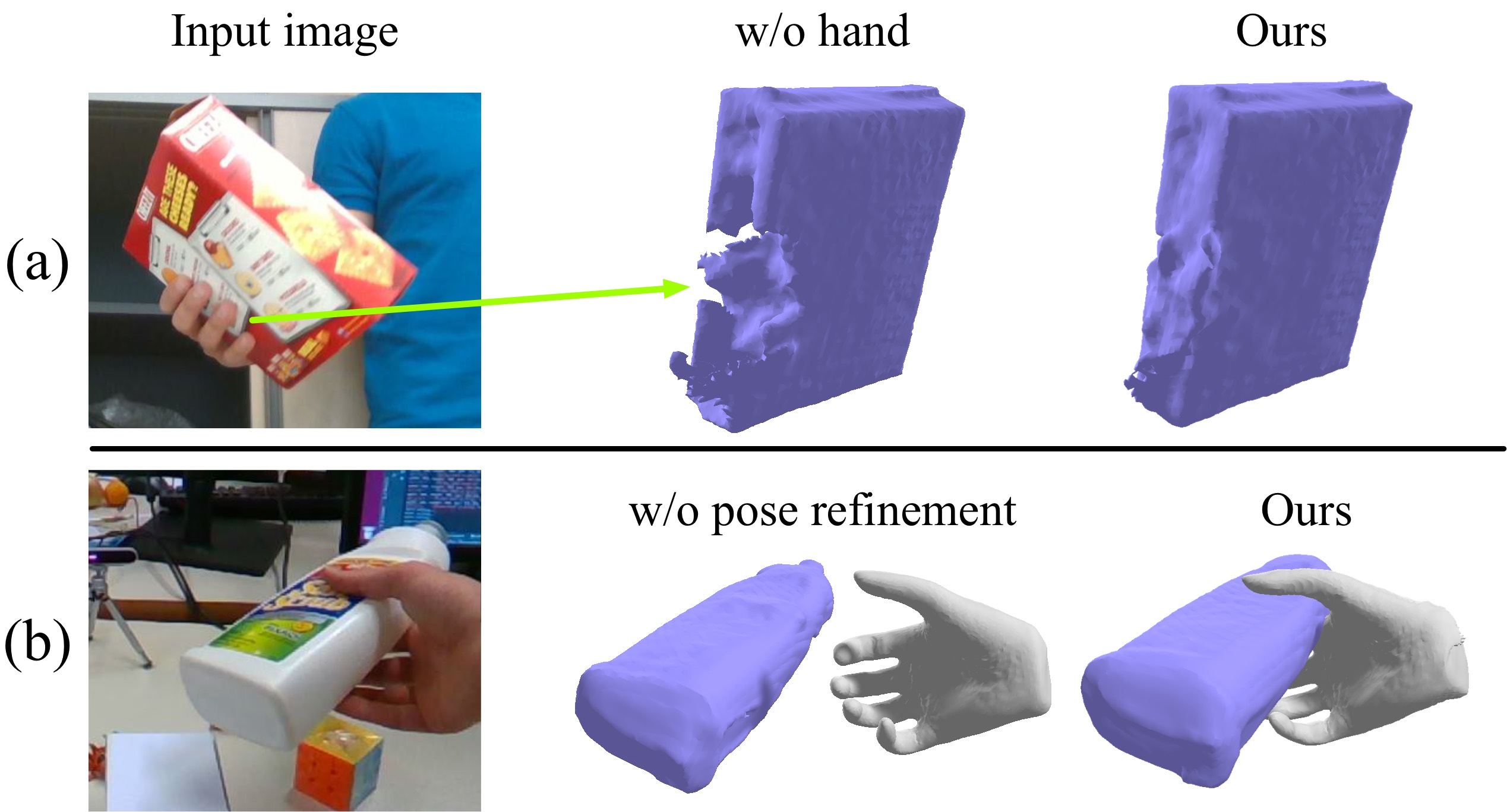}}
    \caption{
        \subtitle{Ablation study}
        \emph{(a)} Jointly reconstructing hand and object effectively reduces artifacts. \emph{(b)} Without contact-based pose refinement, the hand and object can have an erroneous spatial arrangement due to depth ambiguity. 
    }
    \label{fig:ablation}
\end{figure}

\myparagraph{Joint hand-object reconstruction}
To verify that hand reconstruction is complementary to object reconstruction, we implement an ablative baseline without modeling the hand. To be specific, we mask out the hand from all video frames and train the object network on these processed frames. As demonstrated in \refTab{tab:ablation}, removing the hand from our model leads to degraded reconstruction accuracy (CD and F10). A qualitative example is shown in \reffig{fig:ablation} (a). Without hand modeling, the reconstructed object has a hole at the hand-grasping region because the object model needs to fit the images with hand masked out. By jointly modeling the hand, the object, and their occlusion, our method can faithfully reconstruct the object despite hand-object occlusion.

\myparagraph{Contact-based hand-object pose refinement}
To assess the impact of pose refinement as described in Section~\ref{sec:pose_refine}, we compare our model to a baseline that omits this process. \refFig{fig:ablation}(b) provides a rotated-view illustration that highlights the disparity between the baseline model and our full approach. Without pose refinement, there is an unrealistic separation between the hand and object, a common issue in monocular reconstructions due to significant depth ambiguity leading to spatial misalignments.

Our refinement strategy mitigates this by encouraging hand-object contact, thereby diminishing the relative depth uncertainty. The improvements in pose accuracy for both the hand and the object, as well as their spatial arrangement, are quantitatively evidenced in~\reftab{tab:ablation}. Our method outperforms the baseline by achieving superior hand pose accuracy, indicated by lower MPJPE, and improved relative hand Chamfer distance (CD$_h$). These improvements in pose accuracy also translate into more accurate object reconstructions, as reflected by our reduced Chamfer Distance (CD) and F10 scores.

\section{Conclusion}
\label{sec:conclusion}
In this paper, we propose \methodname\ -- the first category-agnostic method that reconstructs an articulated hand and object jointly from a monocular interaction video.
We present a novel compositional implicit model of the object and articulated hand that disentangles and reconstructs 3D hands and objects from 2D observations.
We further show that jointly optimizing the hand and object via interaction constraints leads to better  reconstruction of   object surfaces than reconstructing objects in isolation.
Our method significantly outperforms fully-supervised \sota baselines in both in-the-lab and in-the-wild settings while not relying on \threeD hand-object annotation data. 
Furthermore, we qualitatively demonstrate our method's robustness on challenging in-the-wild videos. 

\myparagraph{Limitations and discussion}
While our method successfully reconstructs hand-object interactions without specific object templates, it does face some challenges. The reconstruction of thin or textureless objects is limited by our use of detector-based Structure from Motion for pose initialization. Advances in detector-free SfM (\eg,~\cite{he2023dfsfm,sun2021loftr})  could potentially address this issue. Furthermore, our reliance on raw \rgb data for supervision may hinder the reconstruction of rarely observed object regions. This could improve with the integration of diffusion priors ~\cite{poole2022dreamfusion} for better object region regularization.

\noindent\textbf{Disclosure}. MJB has received research gift funds from Adobe, Intel, Nvidia, Meta/Facebook, and Amazon.  MJB has financial interests in Amazon, Datagen Technologies, and Meshcapade GmbH.  While MJB is a consultant for Meshcapade, his research in this project was performed solely at, and funded solely by, the Max Planck Society.
{
    \small
    \bibliographystyle{ieeenat_fullname}
    \bibliography{main}
}

\end{document}